\documentclass{article}

\usepackage[final]{nips_2018}

\usepackage[utf8]{inputenc} 
\usepackage[T1]{fontenc}    
\usepackage{hyperref}       
\usepackage{url}            
\usepackage{booktabs}       
\usepackage{amsfonts}       
\usepackage{nicefrac}       
\usepackage{microtype}      
\usepackage{graphicx}\graphicspath{{figures/}}
\usepackage{textcomp}
\usepackage{siunitx} 
\usepackage{sidecap}\sidecaptionvpos{figure}{b}
\usepackage{multirow}

\title{On the Transferability of Representations in Neural Networks Between Datasets and Tasks}

\author{
  Haytham M.~Fayek~~~~~~Lawrence Cavedon~~~~~~Hong Ren Wu \\
  RMIT University \\
  \texttt{haytham.fayek@ieee.org, \{lawrence.cavedon, henry.wu\}@rmit.edu.au}  \\
}

\begin{document}

\maketitle

\begin{abstract}
  Deep networks, composed of multiple layers of hierarchical distributed representations, tend to learn low-level features in initial layers and transition to high-level features towards final layers.
  Paradigms such as transfer learning, multi-task learning, and continual learning leverage this notion of generic hierarchical distributed representations to share knowledge across datasets and tasks.
  Herein, we study the layer-wise transferability of representations in deep networks across a few datasets and tasks and note some interesting empirical observations.
\end{abstract}


\section{Introduction}

Deep networks, composed of multiple layers of hierarchical distributed representations, tend to learn low-level features in initial layers and transition to high-level features towards final layers~\citep{Zeiler2014}.
Similar low-level features commonly appear across various datasets and tasks, while high-level features are somewhat more attuned to the dataset or task at hand, which makes low-level features more generic and easier to transfer from one dataset or task to another~\citep{Yosinski2014}.

Paradigms such as transfer learning~\citep{Pan2010, Bengio2012}, multi-task learning~\citep{Caruana1997, Misra2016}, and continual learning~\citep{Li2016, Rusu2016} leverage this notion of generic hierarchical distributed representations to share knowledge across datasets and tasks.
For example, in transfer learning, typically when data in the target task is scarce, the transfer of low-level features from one dataset or task to another, followed by learning high-level features, is likely to lead to a boost in performance given that both datasets or tasks share some similarity~\citep{Razavian2014}.
Conversely, transferring high-level features and learning low-level ones can be regarded as a form of domain adaptation and can be useful when the tasks are similar or identical but the data distributions are slightly different~\citep{Glorot2011, Bengio2012}.

Herein, we study the layer-wise transferability of representations in deep networks across a few datasets and tasks and note some interesting empirical observations.
First, the layer-wise transferability between two datasets or tasks can be non-symmetric, i.e., features learned for a primary dataset or task can be more relevant to a secondary dataset or task compared with the relevance of features learned for the secondary dataset or task to the primary one, despite both datasets being of similar size.
Second, the nature of the datasets or tasks involved and their relationship is more influential on the layer-wise transferability of representations compared with other factors such as the architecture of the neural network.
Third, the layer-wise transferability of representations can be used as a proxy for quantifying task relatedness.
These observations highlight the importance of curriculum methods and structured approaches to designing systems for multiple tasks in the above mentioned paradigms that can maximize the knowledge transfer and minimize the interference between datasets or tasks.

Layer-wise transferability of representations in deep networks was studied in several prior studies, e.g.,~\citep{Yosinski2014,Fayek2016a, Misra2016}.
In~\citep{Yosinski2014}, the transferability of learned features in a Convolutional Neural Network (ConvNet) trained for an image recognition task was experimentally studied,
where the specificity versus generality of each layer in the ConvNet was quantified using curated classes from the ImageNet dataset.
It was shown that initial layers in deep networks were more transferable than final layers.
A similar study was carried out in~\citep{Misra2016} reporting similar findings.
This work differs from the studies in~\citep{Yosinski2014, Fayek2016a, Misra2016} in that we study the layer-wise transferability between more than just two image recognition datasets, i.e., CIFAR-10, CIFAR-100~\citep{Krizhevsky2009}, and SVHN~\citep{Netzer2011}, and moreover, we study the layer-wise transferability between two speech recognition tasks, i.e., Automatic Speech Recognition (ASR) using the TIMIT dataset~\citep{Garofolo1993} and Speech Emotion Recognition (SER) using the IEMOCAP dataset~\citep{Busso2008}, using more than a single ConvNet architecture, which can provide insights into the influence of neural network architectures on the transferability of representations.


\section{Gradual Transfer Learning}

\begin{figure}
  \centering
  \includegraphics[width=0.95\linewidth]{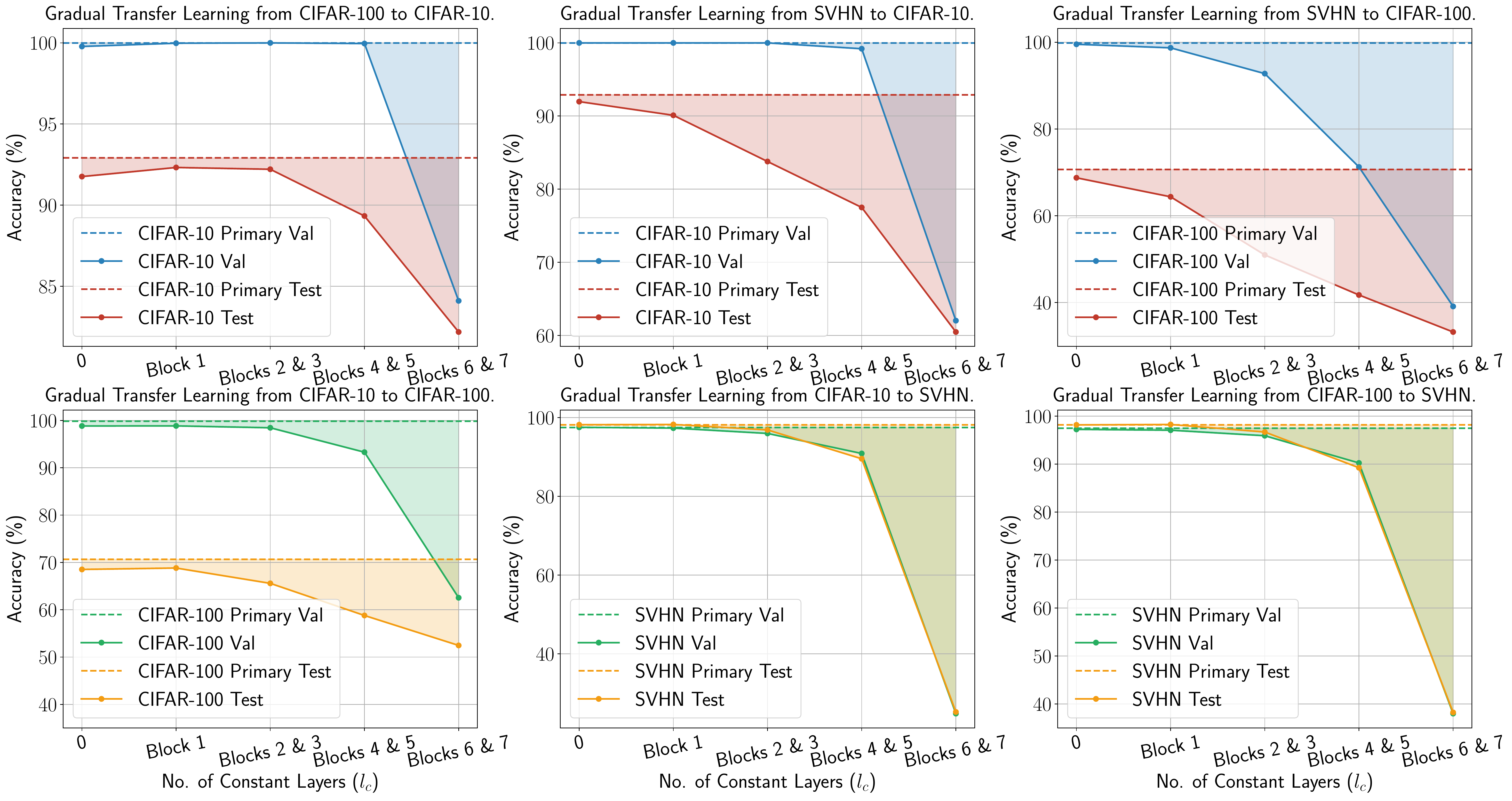}
  \caption{Classification accuracy of gradual transfer learning between the CIFAR-10, CIFAR-100, and SVHN datasets.}%
  \label{fig:image}
\end{figure}

The methodology for quantifying the layer-wise transferability of representations between two datasets or tasks, denoted \textit{gradual transfer learning}, is as follows.
First, two primary neural network models that comprise \( L \) layers are trained for each dataset or task independently.
Second, for each of the two primary models, the learned parameters in all layers of the trained model, except the output layer, are copied to a new model for the (other) secondary dataset or task;
the output layer can be randomly initialized, since it is closely tied to the dataset or task at hand, e.g., the number of output classes in both datasets or tasks may be different.
Third, the first \( l_c \in \{0, \ldots, L_H\} \) layers are held constant and the remaining layers are fine-tuned for the secondary dataset or task, where \( L_H \) is the number of hidden layers in the model, i.e., \( L_H = L - 1 \).
If the constant transferred layers \( l_c \) are relevant to the secondary dataset or task, one can expect an insignificant or no drop in performance relative to the primary model trained independently, and vice versa.
By iteratively varying the number of constant layers \( l_c \), the layer-wise transferability of representations learned for each dataset or task to the other can be inferred.

Iterating \( l_c \) through \( \{0, \ldots, L_H\} \) yields a number of special cases as follows.
In the case of \( l_c = L_H \), the primary model can be regarded as a feature extractor to the secondary model, in that the output layer is the only layer to be fine-tuned.
In the case of \( 1 \leq l_c < L_H \), the output layer is first fine-tuned for a small number of iterations to avoid back-propagating gradients from randomly initialized parameters to previous layers when the output layer is randomly initialized, and subsequently the final \( (L - l_c) \) layers (including the output layer) are fine-tuned simultaneously.
In the case of \(l_c = 0\), the output layer is first fine-tuned for a small number of iterations, and then all layers of the model are fine-tuned simultaneously with the output layer;
in this case, the primary model can be regarded as only an initialization to the secondary model.


\section{Experiments and Results}

\begin{SCfigure}
  \centering
  \includegraphics[width=0.62\linewidth]{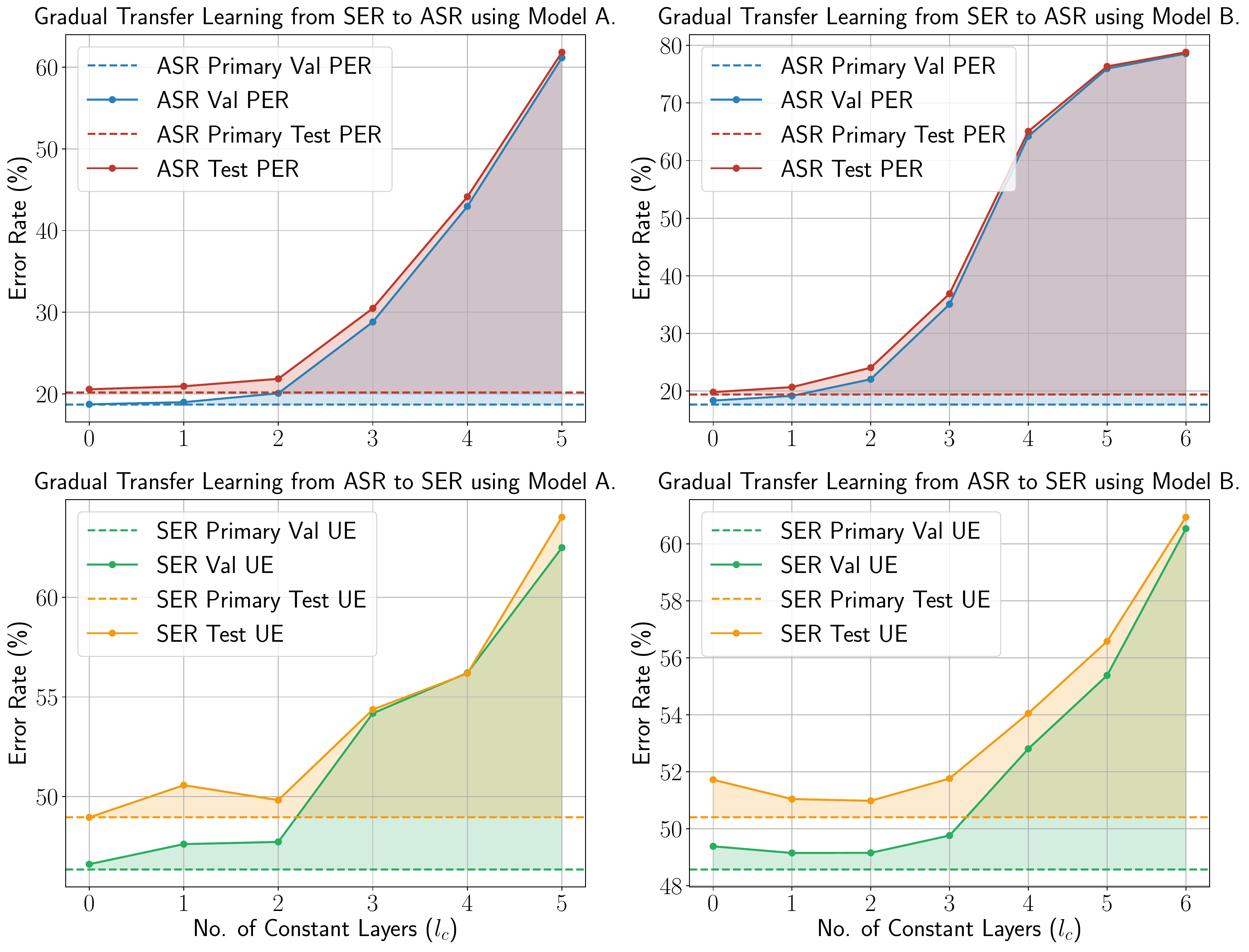}
  \caption{Phone Error Rate (PER) and Unweighted Error (UE) of gradual transfer learning between the ASR (TIMIT) and SER (IEMOCAP) tasks. The results of Model A appeared in~\cite{Fayek2016a}.}%
  \label{fig:speech}
\end{SCfigure}

\paragraph{Layer-wise transferability between the CIFAR-10, CIFAR-100, and SVHN datasets.} 
The CIFAR-10, CIFAR-100, and SVHN datasets are chosen to study how task relatedness can influence the layer-wise transferability.
The CIFAR-10 and CIFAR-100 datasets are both natural images labelled into \( 10 \) and \( 100 \) classes respectively, whereas, the SVHN dataset is street view images of house numbers labelled into \( 10 \) classes corresponding to the \( 10 \) digits;
i.e., it can be expected that the CIFAR-10 and CIFAR-100 datasets are more closely related to each other compared with the SVHN dataset.
For each CIFAR dataset, the original training set was split into a training set of \( 45000 \) images and a validation set of \( 5000 \) images; the entire test set was used for testing.
For the SVHN dataset, the original training set and additional set were combined and split into a training set of \( 598388 \) images and a validation set of \( 6000 \) images; the entire test set was used for testing.
Standard dataset pre-preprocessing was applied to all datasets~\citep{Goodfellow2013, Long2015}, i.e., the mean and standard deviation of the images in the CIFAR datasets were normalized to zero and one respectively using the mean and standard deviation computed from the training set, while, the images in the SVHN dataset were divided by \( 256 \) to lie in the \( [0,1] \) range.

The model used in this experiment follows the Densely Connected Convolutional Networks (DenseNet) architecture~\citep{Huang2017} that comprises \( 40 \) layers (see supplementary material for more details).
It was shown to achieve state-of-the-art performance on the datasets used in this experiment~\citep{Huang2017}.
The main layers in the architecture can be grouped into blocks based on their type and role.
The first block, Block \( 1 \), is a standard convolutional layer.
The dense blocks, Blocks \( 2 \), \( 4 \), and \( 6 \), comprise \( 12 \) layers of Batch Normalization (BatchNorm), Rectified Linear Units (ReLUs), convolution, and dropout.
Each convolution layer in Blocks \( 2 \), \( 4 \), and \( 6 \) is connected to all subsequent layers in the same block via the concatenation operation.
The transition blocks, Blocks \( 3 \) and \( 5 \), are used to counteract the growth in the number of parameters due to the use of the concatenation operation, and are composed of a layer of BatchNorm, ReLUs, convolution, dropout, and average pooling.
A down-sampling block, Block \( 7 \), is used to further reduce the complexity of the model, and is composed of BatchNorm, ReLUs, and average pooling.
The output layer is a fully connected layer followed by a softmax function.
The models were trained following the settings detailed in~\citep{Huang2017}. 
Three primary models were trained independently for the CIFAR-10, CIFAR-100, and SVHN datasets.
Gradual transfer learning was used to assess the layer-wise transferability of the learned representations for each dataset to the other two.
Due to the large number of layers in the model, the number of fixed layers was varied in block intervals as opposed to single layer intervals, i.e., \( l_c \in \{0, Block~1,\allowbreak Blocks~2~and~3,\allowbreak Blocks~4~and~5,\allowbreak Blocks~6~and~7\} \).

The results of gradual transfer learning between the CIFAR-10, CIFAR-100, and SVHN datasets are plotted in \figurename~\ref{fig:image}.
It is observed that the representations learned in the CIFAR-10 and CIFAR-100 datasets are more transferable, i.e., lead to a smaller degradation in performance compared with the primary model, to all other datasets compared with the representations learned in the SVHN dataset.
The learned representations in the SVHN dataset were less transferable to the CIFAR-10 and CIFAR-100 datasets suggesting that the layer-wise transferability of learned representations can be non-symmetric, and moreover, dependant on the nature of the primary and secondary datasets or tasks.
Note that the classes in the CIFAR datasets are more general than the classes in the SVHN dataset that correspond to digits \( 0 \) to \( 9 \).

\paragraph{Layer-wise transferability between the TIMIT and IEMOCAP datasets.} 
Both tasks, the ASR task and the SER task, are speech recognition tasks, yet the relatedness between both tasks is somewhat fuzzy (see~\citep{Fayek2016a} for more details).
For the TIMIT dataset, the complete \( 462 \)-speaker training set, without the dialect (SA) utterances, was used as the training set; the \( 50 \)-speaker development set was used as the validation set; the \( 24 \)-speaker core test set was used as the test set.
For the IEMOCAP dataset, utterances that bore only the following four emotions: \textit{anger}, \textit{happiness}, \textit{sadness}, and \textit{neutral}, were used, with \textit{excitement} considered as \textit{happiness}, amounting to a total of \( 5531 \) utterances.
An eight-fold leave-one-speaker-out cross-validation scheme was employed in all experiments using eight speakers, while the remaining two speakers were used as a validation set.
The results in this experiment are the average of the eight-fold cross-validation.
For both datasets, utterances were split into \SI{25}{ms} frames with a stride of \SI{10}{ms}, and a Hamming window was applied, then \( 40 \) \( \log \)-MFSCs were extracted from each frame.
The mean and standard deviation were normalized per coefficient to zero and one respectively using the mean and standard deviation computed from the training set only in the case of ASR and from training subset in each fold in the case of SER.\@

The ASR system had a hybrid ConvNet-HMM architecture.
A ConvNet acoustic model was used to produce a probability distribution over the states of three-state HMMs with a bi-gram language model estimated from the training set.
The SER system comprised only a ConvNet acoustic model identical to the model used in ASR.\@
Two ConvNet architectures were used to isolate architecture-specific behaviour and trends.
The first architecture, denoted Model \textbf{A}, is a standard ConvNet that comprises two convolutional and max pooling layers, followed by four fully connected layers, with BatchNorm and ReLUs interspersed in-between (see supplementary material for more details).
%
The second architecture, denoted Model \textbf{B}, is a variant of the popular VGGNet architecture~\citep{Sercu2016}.
The architecture comprises a number of convolutional, BatchNorm, and ReLUs layers, with a few max pooling layers used throughout the architecture (see supplementary material for more details),
followed by three fully connected layers, with BatchNorm, ReLUs, and dropout interspersed in-between.
In both architectures, the final fully connected layer is followed by a softmax function to predict the probability distribution over \( 144 \) classes in the case of ASR, i.e., three HMM states per \( 48 \) phonemes, or \( 5 \) classes in the case of SER.\@
The models were trained following the settings detailed in~\citep{Fayek2016a}. 
Two primary models were trained independently for the TIMIT and IEMOCAP datasets for each architecture.
Gradual transfer learning was used to assess the layer-wise transferability of the learned representations for each dataset to the other one.

The results of gradual transfer learning between the TIMIT and IEMOCAP datasets are plotted in \figurename~\ref{fig:speech}.
It is observed that both architectures exhibit similar behaviour, where initial layers are more transferable than final layers and the transferability decreases gradually as we traverse deeper into the network.
With the exception of a few outliers, the layer-wise transferability was similar for both architectures, despite the differences between both architectures in  the number and type of layers. 


\section{Discussion}

The layer-wise transferability of representations was explored on a variety of neural network architectures, datasets, and tasks.
As demonstrated in \figurename~\ref{fig:image}, the layer-wise transferability between two datasets or tasks can be non-symmetric, where the representations learned in the CIFAR-10 and CIFAR-100 datasets were found to be more transferable than the representations learned in the SVHN dataset.
This reflects the nature of the classes in the CIFAR datasets, which are more general, compared with the classes in the SVHN dataset, which correspond to digits \( 0 \) to \( 9 \).
As demonstrated in \figurename~\ref{fig:speech}, the nature of the datasets or tasks involved and their relationship is more influential on the transferability of representations compared with the architecture of the neural network.
Generally, consistent behaviour emerged reflecting the nature of the datasets or tasks involved.
These observations highlight the importance of curriculum methods and structured approaches to designing systems for multiple tasks in paradigms that incorporate learning multiple tasks to maximize the knowledge transfer and minimize the interference between datasets or tasks.


\subsubsection*{Acknowledgments}

H.~M.~Fayek is funded by the Vice-Chancellor's Ph.D.\ Scholarship (VCPS) from RMIT University.
This research was undertaken with the assistance of resources and services from the National Computational Infrastructure (NCI), which is supported by the Australian Government.
The authors gratefully acknowledge the support of NVIDIA Corporation with the donation of one of the Tesla K40 GPUs used for this research.


\setlength{\bibsep}{5pt plus 0.3ex}
\begin{small}
\vspace{-0.1in}
\bibliographystyle{plainnat}
\bibliography{references}
\end{small}


\pagebreak
\section*{Supplementary Material}

\tablename~\ref{tab:convnetimage} details the ConvNet architecture used in the image recognition experiment.
\tablename~\ref{tab:convnetspeechA} and \tablename~\ref{tab:convnetspeechB} detail respectively the ConvNet architecture for Model \textbf{A} and Model \textbf{B} used in the speech recognition experiment.

\vspace{3\baselineskip}

\begin{table}[!h]
  \renewcommand{\arraystretch}{0.9}
  \caption{Densely connected convolutional network architecture for image recognition.
    The outputs of the convolutional layers in Blocks 2, 4, and 6, are concatenated with the inputs to the layer and fed to the subsequent layer in the same block.\protect\footnotemark[1]}%
  \label{tab:convnetimage}
  \centering
  \small
  \begin{tabular}{c c c c c}
    \toprule
    \textbf{Block}     & \textbf{Repeat}                 & \textbf{Type}   & \textbf{Size}         & \textbf{Other} \\
    \midrule
    1                  & \( 1\times \)                   & Convolution     & \( 24, 3 \times 3 \)  & Stride = 1     \\ \midrule 

    \multirow{4}{*}{2} & \multirow{4}{*}{\( 12\times \)} & BatchNorm       & ---                   & ---            \\
                       &                                 & ReLU            & ---                   & ---            \\
                       &                                 & Convolution     & \( 12, 3 \times 3 \)  & Stride = 1     \\ 
                       &                                 & Dropout         & ---                   & \(r = 0.2\)    \\ \midrule

    \multirow{5}{*}{3} & \multirow{5}{*}{\( 1\times \)}  & BatchNorm       & ---                   & ---            \\
                       &                                 & ReLU            & ---                   & ---            \\
                       &                                 & Convolution     & \( 168, 1 \times 1 \) & Stride = 1     \\ 
                       &                                 & Dropout         & ---                   & \(r = 0.2\)    \\
                       &                                 & Average Pooling & \( 2 \times 2\)       & Stride = 2     \\ \midrule

    \multirow{4}{*}{4} & \multirow{4}{*}{\( 12\times \)} & BatchNorm       & ---                   & ---            \\
                       &                                 & ReLU            & ---                   & ---            \\
                       &                                 & Convolution     & \( 12, 3 \times 3 \)  & Stride = 1     \\ 
                       &                                 & Dropout         & ---                   & \(r = 0.2\)    \\ \midrule

    \multirow{5}{*}{5} & \multirow{5}{*}{\( 1\times \)}  & BatchNorm       & ---                   & ---            \\
                       &                                 & ReLU            & ---                   & ---            \\
                       &                                 & Convolution     & \( 312, 1 \times 1 \) & Stride = 1     \\ 
                       &                                 & Dropout         & ---                   & \(r = 0.2\)    \\
                       &                                 & Average Pooling & \( 2 \times 2\)       & Stride = 2     \\ \midrule

    \multirow{4}{*}{6} & \multirow{4}{*}{\( 12\times \)} & BatchNorm       & ---                   & ---            \\
                       &                                 & ReLU            & ---                   & ---            \\
                       &                                 & Convolution     & \( 12, 3 \times 3 \)  & Stride = 1     \\ 
                       &                                 & Dropout         & ---                   & \(r = 0.2\)    \\ \midrule

    \multirow{3}{*}{7} & \multirow{3}{*}{\( 1\times \)}  & BatchNorm       & ---                   & ---            \\
                       &                                 & ReLU            & ---                   & ---            \\
                       &                                 & Average Pooling & \( 8 \times 8 \)      & Stride = 8     \\ \midrule

    \multirow{2}{*}{8} & \multirow{2}{*}{\( 1\times \)}  & Fully Connected & \( K \)               & ---            \\
                       &                                 & Softmax         & ---                   & ---            \\
    \bottomrule
  \end{tabular}
\end{table}

\footnotetext[1]{\( K \) denotes the number of output classes.}

\makeatletter
\setlength{\@fptop}{0pt}
\makeatother

\begin{table}[!htb]
    \renewcommand{\arraystretch}{0.9}
    \caption{Speech recognition convolutional neural network Model \textbf{A} architecture.\protect\footnotemark[1]}%
    \label{tab:convnetspeechA}
    \centering
    \small
    \begin{tabular}{c c c c}
      \toprule
      \textbf{No.}       & \textbf{Type}   & \textbf{Size}        & \textbf{Other} \\
      \midrule
      \multirow{4}{*}{1} & Convolution     & \( 64, 5 \times 4\)  & Stride = 1     \\ 
                         & BatchNorm       & ---                  & ---            \\
                         & ReLU            & ---                  & ---            \\
                         & Max Pooling     & \( 2 \times 2\)      & Stride = 2     \\ \midrule
      \multirow{4}{*}{2} & Convolution     & \( 128, 3 \times 3\) & Stride = 1     \\ 
                         & BatchNorm       & ---                  & ---            \\
                         & ReLU            & ---                  & ---            \\
                         & Max Pooling     & \( 2 \times 2\)      & Stride = 2     \\ \midrule
      \multirow{3}{*}{3} & Fully Connected & \( 1024\)            & ---            \\
                         & BatchNorm       & ---                  & ---            \\
                         & ReLU            & ---                  & ---            \\
                         & Dropout         & ---                  & \(r = 0.4\)    \\ \midrule
      \multirow{3}{*}{4} & Fully Connected & \( 1024\)            & ---            \\
                         & Batch Norm      & ---                  & ---            \\
                         & ReLU            & ---                  & ---            \\
                         & Dropout         & ---                  & \(r = 0.4\)    \\ \midrule
      \multirow{3}{*}{5} & Fully Connected & \( 1024\)            & ---            \\
                         & BatchNorm       & ---                  & ---            \\
                         & ReLU            & ---                  & ---            \\
                         & Dropout         & ---                  & \(r = 0.4\)    \\ \midrule
      \multirow{2}{*}{6} & Fully Connected & \( K \)              & ---            \\
                         & Softmax         & ---                  & ---            \\
      \bottomrule
    \end{tabular}
  \vspace{3\baselineskip}
    \renewcommand{\arraystretch}{0.9}
    \caption{Speech recognition convolutional neural network Model \textbf{B} architecture.\protect\footnotemark[1]}%
    \label{tab:convnetspeechB}
    \centering
    \small
    \begin{tabular}{c c c c c}
      \toprule
      \textbf{No.}       & \textbf{Repeat}                & \textbf{Type}   & \textbf{Size}        & \textbf{Other} \\
      \midrule
      \multirow{7}{*}{1} &                                & Convolution     & \( 64, 6 \times 5\)  & Stride = 1     \\
                         & \(1\times \)                   & BatchNorm       & ---                  & ---            \\
                         &                                & ReLU            & ---                  & ---            \\ \cmidrule{2-5}
                         &                                & Convolution     & \( 64, 3 \times 3\)  & Stride = 1     \\
                         & \(1\times \)                   & BatchNorm       & ---                  & ---            \\
                         &                                & ReLU            & ---                  & ---            \\ \cmidrule{2-5}
                         & \(1\times \)                   & Max Pooling     & \( 2 \times 2\)      & Stride = 2     \\ \midrule

      \multirow{4}{*}{2} & \multirow{3}{*}{\(2\times \)}  & Convolution     & \( 128, 3 \times 3\) & Stride = 1     \\
                         &                                & BatchNorm       & ---                  & ---            \\
                         &                                & ReLU            & ---                  & ---            \\ \cmidrule{2-5}
                         & \(1\times \)                   & Max Pooling     & \( 2 \times 2\)      & Stride = 2     \\ \midrule

      \multirow{4}{*}{3} & \multirow{3}{*}{\(3\times \)}  & Convolution     & \( 256, 3 \times 3\) & Stride = 1     \\
                         &                                & BatchNorm       & ---                  & ---            \\
                         &                                & ReLU            & ---                  & ---            \\ \cmidrule{2-5}
                         & \(1\times \)                   & Max Pooling     & \( 2 \times 2\)      & Stride = 2     \\ \midrule

      \multirow{4}{*}{4} & \multirow{3}{*}{\(3\times \)}  & Convolution     & \( 256, 3 \times 3\) & Stride = 1     \\
                         &                                & BatchNorm       & ---                  & ---            \\
                         &                                & ReLU            & ---                  & ---            \\ \cmidrule{2-5}
                         & \(1\times \)                   & Max Pooling     & \( 2 \times 2\)      & Stride = 2     \\ \midrule

      \multirow{4}{*}{5} & \multirow{4}{*}{\(1\times \) } & Fully Connected & \( 1024\)            & ---            \\
                         &                                & BatchNorm       & ---                  & ---            \\
                         &                                & ReLU            & ---                  & ---            \\
                         &                                & Dropout         & ---                  & \(r = 0.4\)    \\ \midrule
      \multirow{4}{*}{6} & \multirow{4}{*}{\(1\times \) } & Fully Connected & \( 1024\)            & ---            \\
                         &                                & BatchNorm       & ---                  & ---            \\
                         &                                & ReLU            & ---                  & ---            \\
                         &                                & Dropout         & ---                  & \(r = 0.4\)    \\ \midrule

      \multirow{2}{*}{7} & \multirow{2}{*}{\(1\times \) } & Fully Connected & \( K \)              & ---            \\
                         &                                & Softmax         & ---                  & ---            \\
      \bottomrule
    \end{tabular}
\end{table}

\end{document}